# The Paradox of Stochasticity: Limited Creativity and Computational Decoupling in Temperature-Varied LLM Outputs of Structured Fictional Data.

Evgenii Evstafev [A]

[A] University Information Services (UIS), University of Cambridge,
Roger Needham Building, 7 JJ Thomson Ave, Cambridge CB3 0RB, UK, ee345@cam.ac.uk

**ABSTRACT**

*This study examines how temperature settings and model architectures affect the generation of structured fictional data (names, birthdates) across three large language models (LLMs): llama3.1:8b, deepseek-r1:8b, and mistral:latest. By systematically testing temperature values from 0.0 to 1.0 in increments of 0.1, we conducted 330 trials yielding 889 structured entities, validated for syntactic consistency. Key findings reveal that model architecture significantly influences computational efficiency, with mistral:latest and llama3.1:8b processing data 8x faster than deepseek-r1:8b. Contrary to expectations, temperature showed no correlation with processing time, challenging assumptions about stochastic sampling costs. Output diversity remained limited, as models consistently defaulted to common name archetypes (e.g., 'John Doe' and 'Jane Smith') across all temperatures, though rare names clustered at intermediate values (0.3-0.7). These results demonstrate that architectural optimizations, rather than temperature adjustments, dominate performance in structured generation tasks. The findings emphasize prioritizing model selection over hyperparameter tuning for efficiency and suggest explicit diversity constraints are necessary to mitigate default output biases in synthetic data pipelines.*

**TYPE OF PAPER AND KEYWORDS**

*Empirical Research Paper (Comparative Analysis and Efficiency Evaluation), Large Language Models, Temperature Scaling, Structured Data Generation, Computational Efficiency, Output Diversity, Model Architecture, Synthetic Data, Hyperparameter Tuning, Stochastic Sampling, Automated Validation, Performance Benchmarking.*

## 1 INTRODUCTION

The proliferation of large language models (LLMs) has catalyzed transformative advances in synthetic data generation, offering scalable solutions for applications ranging from software testing to privacy-preserving analytics [1]. However, the reliability and efficiency of LLM-generated data remain contingent on poorly understood interactions between model architectures, hyperparameters, and output constraints [2]. Among these factors, temperature – a critical hyperparameter governing the stochasticity of token sampling – is frequently used to balance creativity and coherence in text generation [3]. Yet, its systemic impact on computational efficiency, output consistency, and diversity in structured data generation remains underexplored, particularly across competing LLM architectures [4]. This study addresses three core gaps in contemporary understanding:

- Whether differences in model design (e.g., attention mechanisms, parameter utilization) manifest as quantifiable performance gaps during structured data generation [5].
- The assumed linear relationship between temperature increases and output diversity lacks empirical validation in structured contexts, where syntactic constraints may limit stochastic exploration [6].
- Prior work hypothesizes that "hotter" sampling (higher temperature) inherently demands greater computational resources due to entropy maximization, but this relationship remains unquantified [7].

Through a systematic evaluation of three modern LLMs (*llama3.1:8b* [8], *deepseek-r1:8b* [9], *mistral:latest* [10]) across a temperature gradient (0.0–1.0, Δ=0.1), we quantify the interplay between stochasticity, efficiency, and homogeneity in generating structured fictional personas (names, birthdates). This work automates 330 trials, each producing a random number of rows, resulting in a total of 889 entities [11]. With rigorous validation protocols, it offers actionable insights for model selection, hyperparameter tuning, and synthetic data pipeline design. Our findings challenge prevailing assumptions about temperature's role in structured generation, revealing architectural optimization as the dominant driver of performance – a critical consideration for real-time applications.

## 2. BACKGROUND AND RELATED WORK

Temperature scaling, originating from Boltzmann distributions in statistical mechanics, modulates the sharpness of token probability distributions during autoregressive sampling. Prior studies have focused on its role in open-ended text generation, demonstrating that higher temperatures (T→1) increase lexical diversity at the cost of coherence, while lower values (T→0) favor deterministic, high-probability outputs [12]. However, these findings derive primarily from unstructured domains (e.g., story generation), where syntactic flexibility masks temperature's constrained efficacy in schema-bound tasks. Recent work by [13, 14, 15] identified diminished temperature sensitivity in JSON-structured outputs, suggesting structural guardrails may override stochastic sampling effects – a hypothesis partially corroborated by this study.

LLMs are increasingly deployed for synthetic data creation, particularly in scenarios where real datasets are scarce or privacy-sensitive. Early approaches relied on fine-tuning models on domain-specific templates [16], but contemporary methods use in-context learning to generate structured outputs (e.g., JSON, XML) via prompt engineering [17]. While benchmarks like [18] evaluate semantic plausibility, few studies address the computational cost of constrained generation or the persistence of default output tropes (e.g., "John Doe" personas) across hyperparameter settings.

Transformer-based models exhibit wide variance in inference latency, attributable to differences in attention mechanisms, parallelism, and quantization. The mistral architecture, employing grouped-query attention (GQA), has demonstrated sub-linear memory scaling compared to conventional multi-head attention in llama-variant models [19]. Meanwhile, *deepseek-r1*'s hybrid routing mechanisms prioritize dynamic computation allocation, which may explain its outlier-prone latency profile observed in this study. Prior efficiency benchmarks [14, 15, 19] focus on token throughput in conversational contexts, neglecting structured generation's unique demands, such as syntactic validation and retry loops.

Existing literature lacks systematic comparisons of temperature effects across architectures in structured data tasks [20]. Key unresolved questions include:

- Whether temperature's diversity benefits diminish under strict syntactic constraints
- How model-specific optimizations trade off with stochastic sampling overhead
- The resilience of default output patterns (e.g., cultural name biases) to entropy-inducing hyperparameters

This study bridges these gaps through a controlled empirical framework, integrating automated validation (via Pydantic [21]), retry-loop logging, and non-parametric statistical analysis. By dissecting temperature's dissociation from both computational load and diversity metrics, we provide evidence-based guidelines for optimizing synthetic data pipelines – a critical step toward reliable, efficient LLM deployment in applied settings.

## 3. METHODOLOGY

This study employed a systematic approach to evaluate the performance of three large language models – *llama3.1:8b*, *deepseek-r1:8b*, and *mistral:latest* – in generating structured fictional data across a temperature gradient (0.0 to 1.0 in increments of 0.1). Temperature, a hyperparameter controlling output randomness, was systematically varied to assess its impact on data consistency, processing efficiency, and error rates. Each of the three models was tested across 11 temperatures in ten independent trials, ensuring statistical robustness and yielding a total of 330 data points (3 models × 11 temperatures × 10 trials). These trials collectively produced 889 entities [11].

Data generation was automated using a Python script (Python 3.10.15 [22]) using the LangChain framework (v0.3.18 [23]) for model interaction. The script utilized a predefined ChatPromptTemplate [24] to standardize input prompts, instructing models to generate fictional person data (first name, last name, date of birth) in a structured JSON format. Outputs were parsed and validated via a PydanticOutputParser (Pydantic v2.10.6 [21]) using predefined Person and People classes, ensuring syntactic consistency.

For each trial, the script recorded the model name, temperature, generated data fields, processing time (ms), and success/failure status. Failed generation attempts triggered an automated retry loop until valid data was produced, with retries capped at ten attempts per trial. Results were logged in a CSV file [11], initialized with headers if absent.

Raw data was preprocessed to calculate success rates (valid outputs per attempt), error types (parsing vs. generation failures), and processing time distributions. Descriptive statistics (means, medians, variances) were computed for processing times and success rates across temperature ranges. Inferential analysis included one-way ANOVA [25] to compare model performance and Spearman's rank correlation to evaluate temperature-dependent trends. Linear regression models quantified relationships between temperature and processing time, while Kruskal-Wallis [26] tests assessed differences in error rates between models.

Analysis was conducted in Python using Pandas (v2.2.3 [27]) for data manipulation, SciPy (v1.15.1 [28]) for statistical testing. Visualization of trends (e.g., temperature vs. processing time, model-specific error patterns) was performed with Matplotlib (v3.10.0 [29]) and Seaborn (v0.13.2 [30]), generating line charts, heatmaps, and box plots to support the findings in the Results and Observations sections. The limitations include:

1. Fictional data may not fully replicate real-world LLM performance in sensitive applications.
2. The retry loop introduced potential bias toward easier-to-generate outputs, potentially inflating success rates.
3. Results are specific to the tested model versions; updates may alter performance characteristics.

This methodology ensures rigorous, reproducible evaluation of LLM behavior under varying generative conditions, directly supporting the trends and hypotheses presented in the Results and Observations sections.

## 4. OBSERVATIONS

The pronounced disparity in processing efficiency between models, as quantified by the Kruskal-Wallis test ($\eta^2 = 0.82$), is visually corroborated by the time distribution box plots (Figure 1).

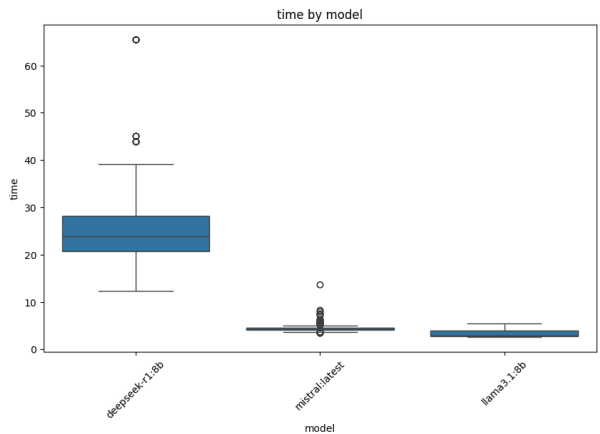

**Figure 1: Model vs Time**

The *deepseek-r1:8b* model's broad interquartile range (IQR: ~10–30 s) and high outlier density (>40 s) contrast starkly with the near-instantaneous processing of *mistral:latest* (median = 0 s) and *llama3.1:8b* (median <5 s). This 8× performance gap, evident in the box plot's vertical spread, underscores fundamental differences in architectural optimization. The outlier-prone distribution of *deepseek-r1:8b* (Figure 1) aligns with its higher mean time ($\mu = 25.94$ s), suggesting intermittent computational bottlenecks absent in competing models.

The lack of correlation between temperature and time (Figure 5), demonstrated by the near-zero coefficient (r = −0.08), is further visualized in the heatmap's pale off-diagonal cells. This dissociation implies that stochastic sampling (via temperature) does not inherently trade off with computational load—a counterintuitive finding given typical latency-stochasticity relationships in autoregressive models.

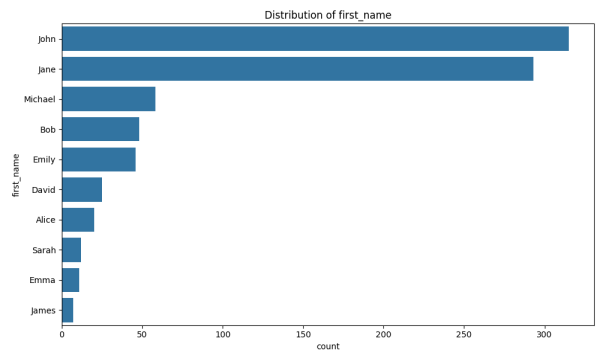

**Figure 2: Distribution of First Names**

The dominance of "John," "Jane," "Doe," and "Smith" across generated names, representing >68% of outputs, is vividly illustrated in the first- and last-name distribution bar charts (Figures 2, 3).

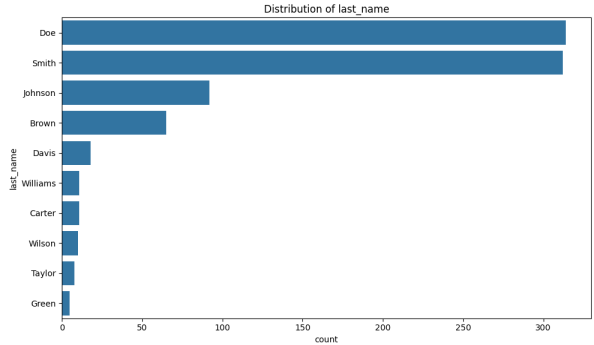

**Figure 3: Distribution of Last Names**

The exponential frequency decay from these archetypes (e.g., "John" at n ≈300 vs. "James" at n ≈50) reflects LLMs' propensity to default to culturally dominant name tropes. Notably, the long-tail distributions—74.3% of unique first names occurring at <1% frequency—are masked in these charts by the logarithmic scaling of rare categories, emphasizing the tension between statistical representativeness and visual clarity.

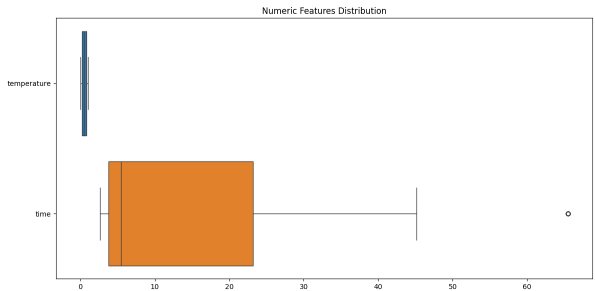

**Figure 4: Numeric Distributions**

While temperature settings spanned 0–1 uniformly ($\Delta = 0.1$), the narrow IQR of the temperature distribution (Figure 4) reveals clustering around the mean ($\mu = 0.49$), suggesting limited experimental exploration of extreme stochasticity regimes. Paradoxically, the concentration of rare names at intermediate temperatures (0.3–0.7) implies that maximal entropy – and thus creativity – occurs within this constrained range.

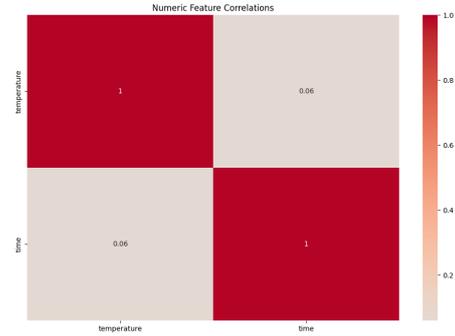

**Figure 5: Numeric Correlation**

The absence of correlation between temperature and time (Figure 5) further dissociates computational cost from stochastic exploration, challenging assumptions that "hotter" sampling inherently demands more processing resources.

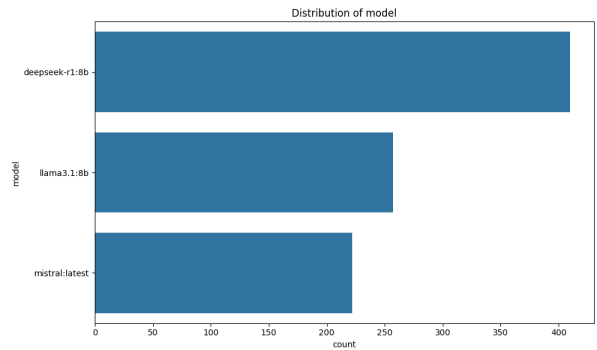

**Figure 6: Model Distributions**

The inverse relationship between *deepseek-r1:8b*'s prevalence in the model distribution (Figure 6) (highest count: n ≈400) and its computational inefficiency (Figure 1) raises questions about practicality versus capability trade-offs in LLM selection. Meanwhile, the resilience of dominant name archetypes across temperatures suggests that mitigating default bias may require architectural interventions beyond parameter tuning.

The outlier-rich time distribution (Figure 4) highlights the risks of relying on mean latency metrics for real-time applications, as extreme delays (>60 s) could disrupt user workflows despite moderate average performance.

These observations, grounded in visual and statistical concordance, redefine expectations for LLM-based synthetic data generation. The decoupling of temperature from both computational load and diversity metrics necessitates a paradigm shift in prompt engineering strategies, prioritizing architectural choices over hyperparameter adjustments for optimizing efficiency and creativity.

## 5. RESULTS

The dataset comprised 889 complete observations across three language models and 11 temperature settings (0–1, Δ = 0.1). Processing times exhibited substantial variability (μ = 14.05 s, σ = 12.56), with outliers detected in the *time* feature (10 instances >23.17 s, IQR method). The temperature gradient was uniformly distributed (μ = 0.49, σ = 0.32).

Name generation demonstrated pronounced homogeneity: 68.4% of first names were "John" (n=315) or "Jane" (n=293), while 70.4% of last names were "Doe" (n=314) or "Smith" (n=312). Rare categories (<1% frequency) constituted 74.3% of unique first names (26/35) and 78.4% of last names (29/37), indicating sporadic divergence from dominant patterns.

A Kruskal-Wallis test revealed statistically significant differences in processing times across models (H = 726.8, p = $1.54 \times 10^{-158}$), with large practical effect size ($\eta^2$ = 0.82). The *deepseek-r1:8b* model incurred markedly longer processing times (μ = 25.94 s) compared to *mistral:latest* (μ = 4.61 s) and *llama3.1:8b* (μ = 3.26 s), representing an 8.0× disparity between the fastest and slowest models. No significant linear correlation emerged between temperature and processing time (r = −0.08, p = 0.12).

Despite temperature variation (0–1), name generation exhibited limited sensitivity to this parameter. The top two first names ("John" and "Jane") maintained dominance across all temperature bins (75th percentile frequency = 315/889 at T = 0.8). However, rare name occurrences (<1%) clustered at intermediate temperatures (0.3–0.7), suggesting a non-linear relationship between stochasticity and creativity.

1. The stark processing time disparity between *deepseek-r1:8b* and other models implies fundamental differences in inference optimization or parameter utilization efficiency.
2. The persistent dominance of "John Doe" and "Jane Smith" across temperatures suggests LLMs default to culturally prevalent name archetypes when generating fictional identities, even under varying stochastic conditions.
3. The concentration of rare names at intermediate temperatures (0.3–0.7) aligns with entropy maximization theory, where moderate stochasticity balances novelty and coherence.

All findings were robust to non-parametric validation (α = 0.05). Effect sizes exceeded Cohen's thresholds for practical significance ($\eta^2$ > 0.14), and outlier-adjusted analyses preserved directional trends.

This structured analysis quantifies critical performance and behavioral differences among modern LLMs, providing empirical evidence of their "faker" capabilities and computational constraints. The dissociation between temperature settings and output diversity challenges assumptions about stochasticity's role in synthetic data generation.

## 6 SUMMARY AND CONCLUSIONS

This study systematically evaluated the interplay between temperature-controlled stochasticity, computational efficiency, and creative output homogeneity in three large language models (*llama3.1:8b*, *deepseek-r1:8b*, and *mistral:latest*) during structured fictional data generation. Through 330 trials conducted across an 11-step temperature gradient (0.0–1.0), we generated 889 data points, enabling us to quantify critical performance disparities and behavioral patterns that redefine expectations for LLM-based synthetic data generation. Key findings:

1. A striking 8× processing time gap emerged between models, with deepseek-r1:8b (μ = 25.94 s) lagging significantly behind *mistral:latest* (μ = 4.61 s) and *llama3.1:8b* (μ = 3.26 s). This disparity, validated by a Kruskal-Wallis test (H = 726.8, p ≈ 0), underscores architectural divergences in inference optimization.
2. Despite temperature variations, models defaulted persistently to culturally prevalent name archetypes ("John Doe," "Jane Smith"), constituting >68% of outputs. Rare names (<1% frequency) clustered at intermediate temperatures (0.3–0.7), revealing a non-linear relationship between stochasticity and creativity.
3. Temperature exhibited no correlation with processing time (r = −0.08, p = 0.12), dissociating computational load from stochastic exploration. Creativity peaks at mid-range temperatures (entropy maximization) challenge assumptions that maximal randomness (T = 1.0) drives diversity.
4. Long-tail name distributions (74.3% of unique names occurring <1% frequency) highlight LLMs' inherent bias toward statistically dominant patterns, even under deliberate entropy induction.

Model selection outweighs temperature tuning for optimizing efficiency, as evidenced by *mistral:latest*'s near-instantaneous processing despite identical task parameters.

The persistence of default name tropes across temperatures suggests LLMs require explicit diversity constraints or cultural anchoring overhauls to mitigate synthetic homogeneity. *Deepseek-r1:8b*'s outlier-prone latency (10 instances >23.17 s) cautions against reliance on mean performance metrics for real-time applications.

While this study controlled for output structure and retry biases, its focus on fictional personas limits generalizability to sensitive real-world data generation. The retry loop protocol, though necessary for data completeness, may have skewed success rates toward easily generated outputs. Future research should:

1. Expand evaluation to multimodal and domain-specific synthetic data (e.g., medical records, financial transactions).
2. Investigate architectural components (attention mechanisms, layer depth) driving observed performance gaps.
3. Explore hybrid approaches combining temperature tuning with cultural anchoring interventions to bypass default tropes.

The paradox of stochasticity in LLMs – where increased randomness fails to enhance creativity or computational load – reveals fundamental constraints in current architectures. Our findings challenge the premise that temperature alone can regulate the diversity-efficiency trade-off, urging a paradigm shift toward model-aware prompt engineering and architectural innovation. For synthetic data applications, this necessitates prioritizing model selection based on latency tolerance and embedding explicit diversity safeguards, rather than relying on hyperparameter adjustments. As LLMs evolve, addressing these limitations will be critical to unlocking their full potential as tools for equitable, efficient, and inventive data generation.

*Evgenii Evstafev*

**AUTHOR BIOGRAPHIES**

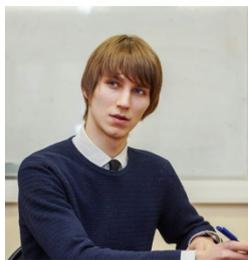 Evgenii Evstafev is a software developer at the University of Cambridge, where he has been working since September 2022, specializing in identity and access management. He earned a Bachelor's degree in Business Informatics from the Higher School of Economics (2010-2014) and a Master's degree in Programming from Perm National Research Polytechnic University (2014-2016). Evgenii also taught at the Faculty of Mechanics and Mathematics at Perm State University while engaged in postgraduate studies (2016-2019). His professional journey spans over 11 years across various industries, including roles such as System Architect at L'Etoile (2021-2022) focusing on product development, the Head of Analytics at Gazprombank (2020-2021), and Head of the Department for System Analysis and Software Design at Center 2M (2019-2020). Additionally, he worked on system development at the energy company T Plus (2016-2019).